\documentclass[10pt]{article}
\usepackage[letterpaper]{geometry}
\usepackage{natbib}
\usepackage{graphicx}
\usepackage{amsthm}
\usepackage{amssymb}
\usepackage{amsmath}
\usepackage{mdwlist}
\usepackage{afterpage}
\usepackage{latexsym,amsmath,amsthm,amsfonts}
\usepackage[section]{placeins}
\usepackage{setspace}
\usepackage{xspace} 
\usepackage{color} 
\usepackage{soul} 
\usepackage{multirow} 
\usepackage{enumerate}

\newcommand{\be}{\begin{equation}}
\newcommand{\ee}{\end{equation}}

\def\real{\mathbb R}

\def\argmax{\mathop{\rm arg\, max}}
\def\argmin{\mathop{\rm arg\, min}}

\newcommand{\bX}{\mbox{$\bf X$}}
\newcommand{\bx}{\mbox{$\bf x$}}
\newcommand{\by}{\mbox{$\bf y$}}
\newcommand{\bz}{\mbox{$\bf z$}}

\newcommand{\T}{^\text{T}}

\graphicspath{{plots/}{plots/paths}}

\usepackage{color}
\usepackage[usenames,dvipsnames]{xcolor}

\begin{document}

\title{\vspace{-0.75in} A Permutation Approach for Selecting the Penalty Parameter \\ in Penalized Model Selection}

\author{
  {\bfseries
  Jeremy Sabourin$^{1}$,
  William Valdar$^{1,2}$,
  and Andrew Nobel$^{3}$ 
  	\thanks{  	
    	Correspondence to:
      	Andrew Nobel, Department of Statistics and Operations Research, University of North Carolina at Chapel Hill, Chapel Hill, NC 27599-7265, USA.
      	E-mail: \texttt{nobel@email.unc.edu}
    }
  }  
  \\
  $^1${\small\itshape Department of Genetics, University of North Carolina at Chapel Hill, North Carolina}
  \\
  $^2${\small\itshape Lineberger Comprehensive Cancer Center, University of North Carolina at Chapel Hill, North Carolina}
  \\
  $^3${\small\itshape Department of Statistics and Operations Research, University of North Carolina at Chapel Hill, North Carolina}  
  \\
}
\date{\today}

\maketitle
\singlespace
\vspace{-0.5in} 
\begin{abstract}
We describe a simple, efficient, permutation based procedure for selecting the penalty parameter 
in the LASSO.  The procedure, which is intended for applications where variable selection is the 
primary focus, can be applied in a variety of structural settings, including generalized linear models.  
We briefly discuss connections between permutation selection and existing theory for the LASSO. 
In addition, we present a simulation study and an analysis of three real data sets in which permutation
selection is compared with cross-validation (CV), the Bayesian information criterion (BIC), and a selection 
method based on recently developed testing procedures for the LASSO.  
\end{abstract}


\doublespace
\section{Introduction}
\label{sec:intro}

The analysis of high dimensional data, in which the number of measured predictors is large and 
can exceed the number of samples, is an important and common problem in statistical applications.   
When samples are accompanied by a real or categorical response, data analysis typically includes model fitting 
with the aim of doing prediction or variable selection, or both.  The goal of prediction is to derive a rule capable 
of accurately predicting the response of a new, unlabeled sample.  
The goal of variable selection is to select a (small) subset 
of the measured predictors whose individual or coordinated activity is significantly related to the response.  
In both cases, it is common to assume that the observed data 
arise from an underlying model that is sparse, in the sense that only a small subset of the predictors are related to the response.  
Whether sparsity is assumed, or viewed as a desirable feature of a model,
analysis of high dimensional data is often carried out by penalized methods that produce 
models in which a relatively
small subset of the available predictors are included.  Popular penalized methods include the 
LASSO \citep{Tibshirani96}, its numerous variations, and SCAD \citep{Fan_Li_2001}.  
In what follows, we focus our attention 
on the LASSO.

The LASSO and its variants require specification of a penalty/tuning parameter that controls the tradeoff between model 
fit and model size.  Specification of this parameter is necessary to fully determine the selected model, and  
each value results in different coefficient estimates; selecting a suitable value of the penalty parameter 
is therefore a vital part of model fitting.  
At present, the most widely used procedures for selecting the penalty parameter in the LASSO are the Bayesian Information Criterion (BIC) and 
cross validation (CV).  Cross validation, which is based on out of sample prediction, is a natural choice when the goal of model fitting is prediction.
CV has some empirical and theoretical justification \citep[{see, for example, B{\"{u}}hlmann's comments in}][]{Tibs11}, but in high-dimensional settings CV tends to be too conservative \citep{Feng2013arXiv}.  
BIC, although designed primarily for variable selection, is based in part on how well the 
selected model fits the data, 
making it a reasonable choice for either prediction or variable selection. 
In practice, BIC is a popular method, but has no theoretical justification for variable selection 
with the LASSO \citep{lassoBook}.

In this paper we address the problem of selecting an appropriate penalty parameter for the LASSO 
when the primary goal of model fitting is variable selection.  We investigate a simple, 
permutation based procedure for choosing the penalty parameter in linear regression and 
generalized linear model settings. 
This permutation procedure was introduced previously in a more limited context in 
\citet{ValdarSabourin12}, and was motivated by work of \citet{Ayers10}.  
The procedure considers the minimal level of penalization required to remove all predictors from the LASSO model
under multiple random permutations of the response.  Permutation of the response provides
a baseline under which there are no true associations with the predictors. 
Applying a comparable level of penalization to the true (unpermuted) response ensures that the variables 
included in the model have a joint relationship with the response that is stronger than joint relationships 
arising by chance.

In recent work, \citet{Lockhart2013} proposed a covariance test for the LASSO that can be applied to 
parameter selection when variable selection is of primary interest.
The test yields a p-value for each model change (inclusion or exclusion of a variable) in the LASSO path.  
For simple linear models, the test requires that the error variance be known
or estimated from the data, limiting general use of the test in this case to
data for which the number of samples is larger than the number of predictors.  
Estimation of the error variance in high dimensional settings, and application of the test to selection of the LASSO 
penalty parameter are currently under investigation by Lockhart et al..  

The remainder of the paper is organized as follows.  In the next section, we outline the general data setting and 
describe the permutation-based selection procedure.  In Section \ref{sec:theory} we draw some connections between
the proposed permutation selection method and the penalty parameters assumed by existing asymptotic 
theory for the LASSO using recent work on maximal correlations of random vectors on the unit sphere.
Section \ref{sec:sims} is devoted to a simulation study in which we compare the proposed permutation method with
CV, BIC, and a selection procedure derived from the test of \citet{Lockhart2013}.
In Section \ref{sec:realdata}, we examine the application of different penalty parameter selection methods to multiple real data sets.
We conclude with a brief discussion in Section \ref{sec:disc}.

\section{Permutation Selection of the Penalty Parameter}
\label{sec:perm}

\subsection{Data Setting and Model}

We assume that the available data is in the form of an an $n \times p$ data matrix $\bX$ and an $n \times 1$ response vector $\by$.
The rows of $\bX$ correspond to samples; the columns $\bx_1, \ldots, \bx_p$ of $\bX$ correspond to measured variables of interest, 
and are assumed to be standardized.  For convenience we describe the proposed permutation scheme in settings where the 
application of a generalized linear model (GLM) is augmented by use of a 1-norm penalty to derive a sparse model from
the available data.  

A GLM consists of three components: a random component associated with the response; 
a systematic component equal to a linear function of the data, 
and a link function connecting the random and systematic components.  
Let $\by = y_1, \ldots, y_n$ be independent observations of a response variable 
$\mathbf{Y}$ whose distribution belongs to an exponential family. 
Here we focus on the Gaussian for continuous responses and on the Binomial for binary, case/control type responses.  Let $\boldsymbol{\mu} = E(\mathbf{Y})$ and let $g(\cdot)$ be a known link function.  
The response and data matrix are linked by the GLM 
\begin{equation}
	\label{eq:glm}
	g(\boldsymbol{\mu}) = \bX \T \boldsymbol{\beta} \,.
\end{equation}

The LASSO procedure \citep{Tibshirani96} fits a sequence, or path, of models characterized by a positive penalty parameter $\lambda$ that trades off between the overall fit of the model and its complexity. 
Let $||\boldsymbol{\beta}||_1 = \sum_{j=1}^p |\beta_j|$ be the 1-norm of $\boldsymbol{\beta}$, and let 
\begin{equation}
\ell(\boldsymbol{\beta} : \by, \bX) = \sum^n_{i=1} \log L(\boldsymbol{\beta} : y_i, \bx_i)\,,
\end{equation}
be the log-likelihood of $\boldsymbol{\beta}$ given $\bX$ and $\by$, 
where $L(\cdot)$ is the likelihood function of a single observation under the GLM (\ref{eq:glm}).
For a specified value of $\lambda \geq 0$ the LASSO procedure identifies the (unique) model defined by the
coefficient vector 
\begin{equation}
\label{eq:lasso}
\hat{\boldsymbol{\beta}} (\lambda : \by, \bX)
 = \underset{\boldsymbol{\beta}}{\argmax} \left\{ \ell(\boldsymbol{\beta}: \by, \bX) - 
\lambda ||\boldsymbol{\beta}||_1 \right\} .
\end{equation}
For each $\lambda \geq 0$ let
\begin{equation}
\label{eq:s0def}
S_0(\lambda) \ = \ \{ j : \hat{\boldsymbol{\beta}}_j(\lambda : \by, \bX) \neq 0 \}
\end{equation}
be the set of variables included in the model associated with the coefficient vector 
$\hat{\boldsymbol{\beta}} (\lambda : \by, \bX)$.
As noted in the introduction, use of the 1-norm penalty ensures that when $\lambda$ is large
the coefficient vector $\hat{\boldsymbol{\beta}} (\lambda : \by, \bX)$ will be sparse, or equivalently, the set $S_0(\lambda)$ of selected variables will be small.
Use of the LASSO and related procedures arises from practical
interest in deriving models of the data that include a relatively small number of selected variables.

\subsection{Permutation Selection Procedure}
\label{sec:perm_def}

For any permutation $\pi$ of $[n] := \{1,\ldots,n\}$ let 
$\by_{\pi} = (y_{\pi(1)}, \ldots, y_{\pi(n)})\T$ be a re-ordered version 
of the response $\by$. 
Suppose that permutations $\pi_1, \ldots, \pi_N$ are obtained by sampling uniformly at 
random from the set of permutations
on $[n]$.  For each $1 \leq l \leq N$ and each $\lambda \geq 0$ let
\[
S_l(\lambda) \ = \ \{j : \hat{\boldsymbol{\beta}}_j(\lambda : \by_{\pi_l}, \bX) \neq 0  \}
\]
be the set of variables selected by the LASSO procedure (\ref{eq:lasso}) with response $\by_{\pi_l}$ and penalty
parameter $\lambda$.  For each random permutation $\pi_l$ let 
\begin{equation}
\label{eq:lambda0}
\lambda_0 (\by_{\boldsymbol{\pi}_l}) 
\ = \ 
\min\left\{ \lambda : | S_l(\lambda) | = 0 \right\} 
\end{equation}
be the smallest value of the penalty parameter for which no variables are selected for the fitted model.  
Permutation of the response ensures that there is, on average, no systematic relation between $\by_{\pi_l}$ and 
the measured predictors in $\bX$.
In this case, exclusion of variables from the fitted model correctly reflects the absence of a relationship between
$\by_{\pi_l}$ and $\bX$.  The quantity $\lambda_0 (\by_{\pi_l})$ is the smallest amount
of penalization for which this null relationship is maintained.  Our proposed 
choice of penalty parameter is the median of the observed minimum penalties, namely,
\begin{equation}
\label{lamhat}
\hat{\lambda}_\text{perm} = \text{median}( \lambda_0(\by_{\pi_1}), \ldots,  \lambda_0(\by_{\pi_N}) ) \, .
\end{equation}
As can be seen from its definition, the choice of $\hat{\lambda}_\text{perm}$ is targeted towards the 
goal of {\em variable selection}; 
the predictive performance of $\hat{\lambda}_\text{perm}$ is considered only briefly
in the real data analyses below.

Variability of $\hat{\lambda}_\text{perm}$, arising from the use of random permutations, 
can potentially lead to variability in the selected model $S_0(\lambda_\text{perm})$. 
Variability of $\hat{\lambda}_\text{perm}$ depends on the number of permutations $N$, as well as
the means by which the values $\lambda_0(\by_{\pi_l})$ are aggregated.
Aggregation using the median provides a relatively stable estimate for low to moderate values of $N$.
\citet{Ayers10} proposed a similar penalty selection procedure 
that uses the maximum of $\lambda_0(\by_{\pi_l})$ over 25 permutations of the response. 
They showed empirically that the procedure controls the 
family-wise error rate of falsely including a variable unrelated to the response.
However, use of the maximum increases variability of the selected model. 
In particular, when evaluating $\lambda$ multiple times on the same data, multiple models can be selected.
To reduce this variability, the number of permutations can be increased, but this has the 
effect of changing the family wise error rate.
In the selection procedure proposed here, increasing the number of permutations  
yields a better estimate of the true median $\lambda_0(\by_\pi)$, and reduces permutation variability.

In our experience with simulated and real data, $N=100$ permutations is sufficient to 
control permutation variability and provide reasonable values of $\hat{\lambda}_\text{perm}$.
In some cases, a lower $N$ can be sufficient: in \citet{ValdarSabourin12} the median of $N=20$ 
permutations was enough to select a stable model $S_0(\hat{\lambda}_\text{perm})$ in a genetics setting.

\section{Some Connections with LASSO Theory}
\label{sec:theory}
In the case of linear regression, recent results on maximal correlations provide an asymptotic connection
between the permutation based selection procedure described here and existing theoretical work on the
consistency of the LASSO.   Assume for the moment that the data matrix $\bX$ and response $\by$ are
related by the standard linear regression model
$
\by \ = \ \bX^T \, \boldsymbol{\beta} + \boldsymbol{\varepsilon},
$
where $\boldsymbol{\varepsilon}$ is a vector of independent mean zero Gaussian errors 
with common variance.
Suppose that $\by$ and the columns
$\bx_1,\ldots,\bx_p$ of $\bX$ have been centered and scaled 
to have mean zero and total sum of squares one; then $\by$ and each $\bx_j$ lie on 
the unit sphere $S^{n-1}$ in $\real^n$.
Recall that the standard LASSO coefficient vector is given by
\begin{equation}
\label{lassor}
\hat{\boldsymbol{\beta}} (\lambda : \by, \bX)
\ = \ \underset{\boldsymbol{\beta}}{\argmin} 
\left\{ \frac{1}{2} \sum_{i=1}^n (\bx_i^T \, \boldsymbol{\beta} - y_i)^2 \, + \, 
\lambda ||\boldsymbol{\beta}||_1 \right\} .
\end{equation}
It is known (cf.\ \citet{Friedman2010}) that the minimal penalty under which all coefficients of
$\boldsymbol{\beta}$ are zero is given by
\begin{equation}
\label{eq:lam0}
\lambda_0(\by) \ = \ \max_{1 \leq j \leq p}  | \bx_j^\text{T} \, \by | \,,
\end{equation}
Thus for any permutation $\pi_l$ of the response $\by$,  
\begin{equation}
\label{pi0m}
\lambda_0(\by_{\pi_l}) \ = \ \max_{1 \leq j \leq p}  | \bx_j^\text{T} \, \by_{\pi_l} | 
\end{equation}
is simply the maximum (unsigned) inner product between $\by_{\pi_l}$ and the columns of $\bX$.

In recent work, \citet{Zhang_arXiv} shows that if both $n$ and $p$ tend to infinity then, 
letting ${\bf U}$ be a random vector uniformly distributed on the unit sphere $S^{n-1}$,  
\begin{equation}
\label{zb1}
\inf_{{\bf v}_1,\ldots, {\bf v}_p \in \real^n} \ 
\mathbb{P} \left( \max_{1 \leq j \leq p}  | {\bf v}_j^\text{T}  {\bf U} | \leq \sqrt{1-p^{-2/(n-1)}}  \right) \ \to \ 1 \, .
\end{equation}
A corresponding lower bound holds if the vectors ${\bf v}_i$ are independent and uniformly distributed
on $S^{n-1}$.  Under the standard assumption that $n$ and $p$ grow in such a way that
$\log p / n \to 0$, it is easy to see that $\sqrt{1-p^{-2/(n-1)}}$ is approximately $\sqrt{2 \log p / n}$, in
the sense that the ratio between these quantities tends to one as $n,p$ tend to infinity.  
Thus for each $\delta > 0$, if $n$ and $p$ are 
sufficiently large and $\log p / n$ is close to zero, then with high probability
\begin{equation}
\label{zb2}
\max_{1 \leq j \leq p}  | \bx_j^\text{T} \, {\bf U} | \ \leq \ (1 + \delta) \sqrt{ \frac{2 \log p}{n}} .
\end{equation}
We expect that a corresponding lower bound, with a different leading constant, will hold for data matrices
whose columns $\bx_j$ are weakly dependent. 
Comparing the last display with equation (\ref{pi0m}), we see that in permutation 
selection the permuted responses
$\by_{\pi_l}$ act as a proxy for a uniform sample from the unit sphere $S^{n-1}$, and therefore
$\hat{\lambda}_\text{perm}$ acts as an estimate of the population quantity
\begin{equation}
\label{median}
\mbox{median} \left( \max_{1 \leq j \leq p}  | \bx_j^\text{T} \, {\bf U} | \right) .  
\end{equation}
The quantity $\sqrt{2 \log p / n}$ 
appearing in (\ref{zb2}) is, up to constants, the value of the penalty parameter assumed in standard 
results (cf.\ \citet{lassoBook}) concerning the asymptotic properties of the LASSO for prediction and
variable selection.  Recent theoretical work 
(c.f.\ \cite{Lederer2013}, \cite{Dalalyan2014}, and the references therein) has examined in some
detail how correlations among the columns $\bx_j$ of the design matrix $\bX$ affect the performance
of the LASSO and the selection of the penalty parameter $\lambda$.  The broad conclusion of this work
is that smaller values of $\lambda$ are appropriate when the columns of $\bX$ are more correlated,
a relationship that holds for the permutation selection procedure.  Indeed, the permutation selection
procedure adapts to linear dependence among the columns of the design matrix in a direct way, by assessing 
their maximum correlation with a permuted response.

\begin{figure}[ht!]
  \centering
	\includegraphics[width=.35\linewidth]{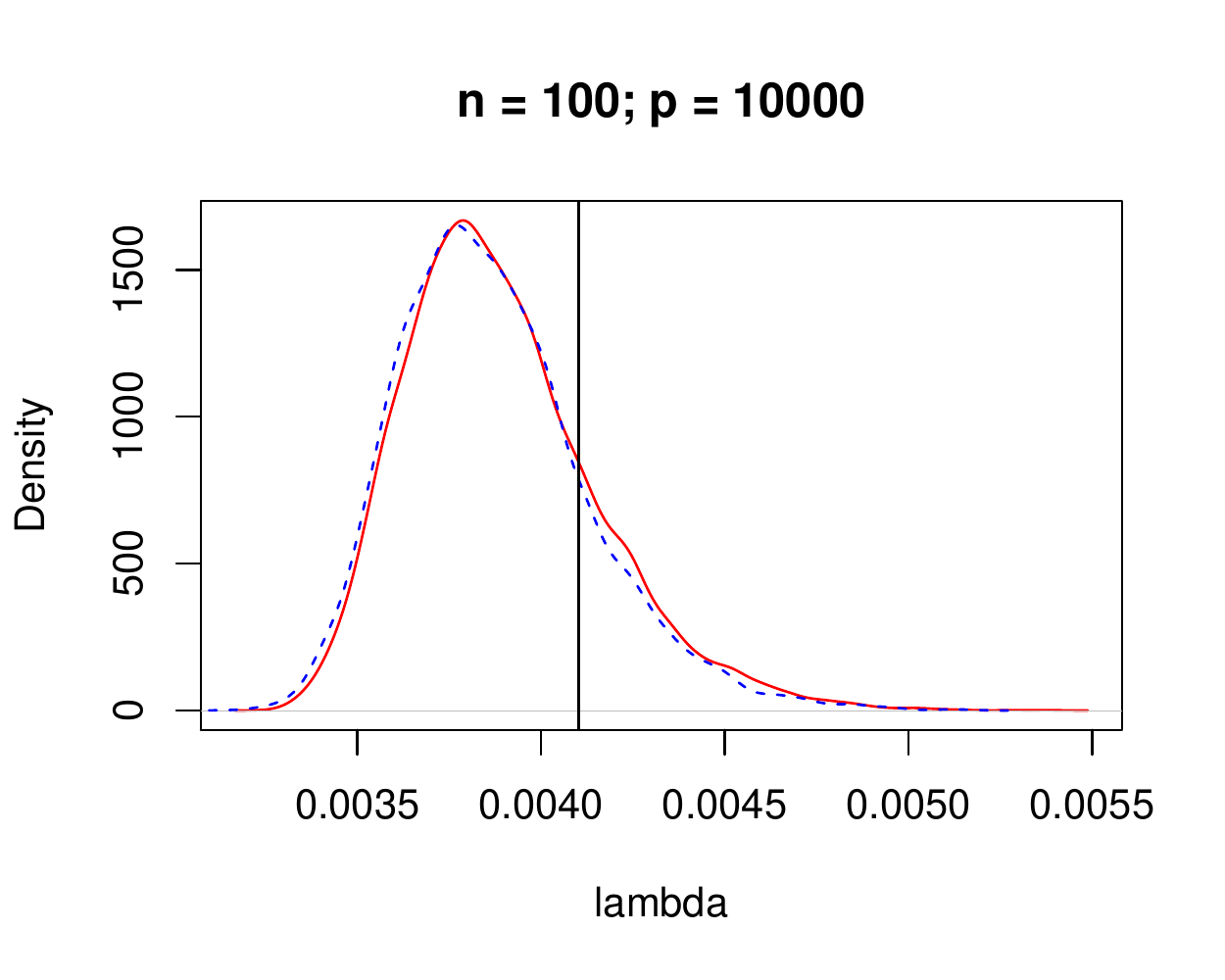}	\includegraphics[width=.35\linewidth]{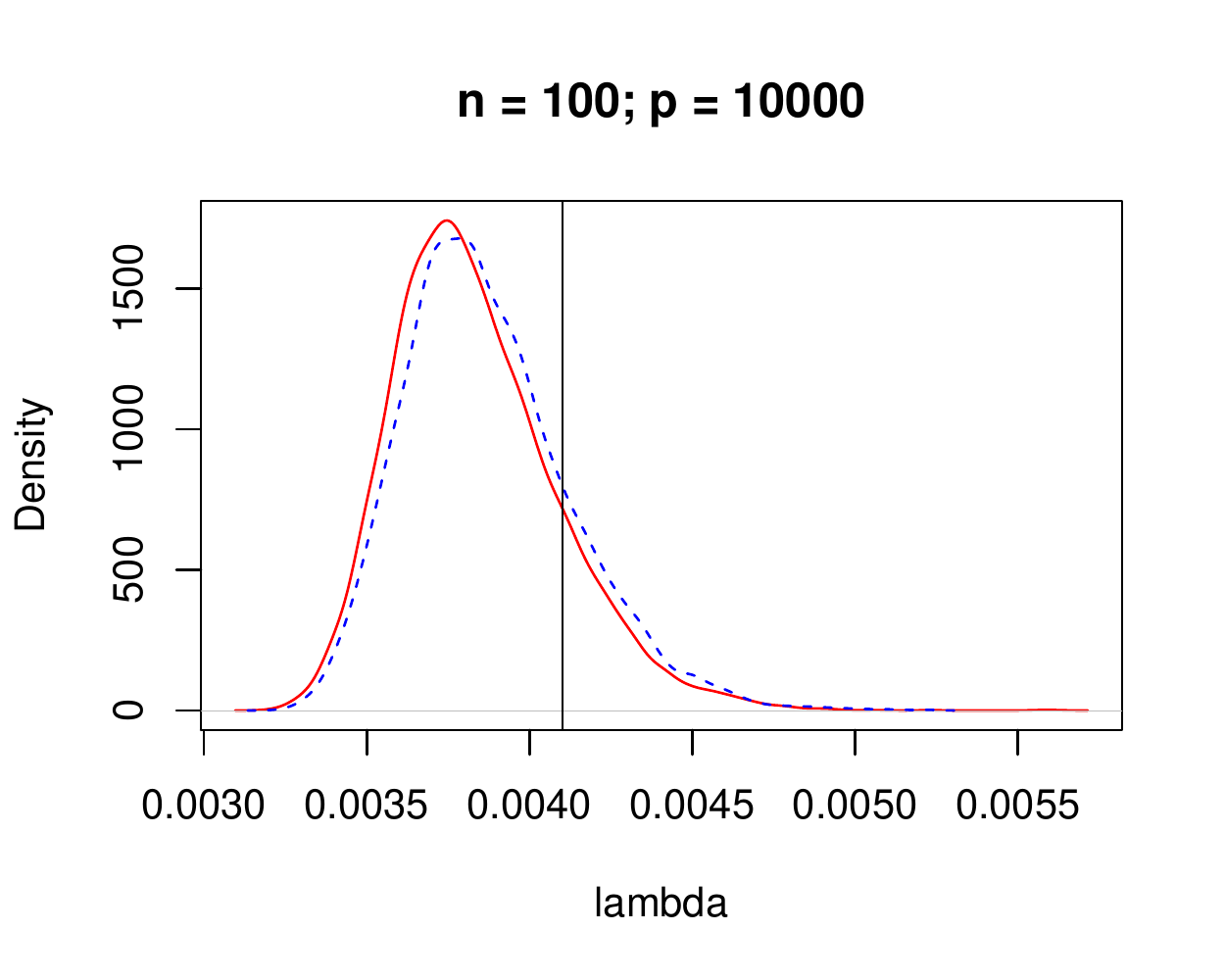}\\
  \caption{Distributions of $\lambda_0(\tilde{\by}_\pi)$ (red) and $\lambda_0(\tilde{\bz})$ (blue) for $n=100$ and $p=10000$.  (Left) sparse model with $s=10$. (Right) non sparse model with $s=1000$. The black vertical line indicates the value of $\sqrt{1-p^{-2/(n-1)}}$. }
  \label{fig:unif}
\end{figure}

To examine the extent to which permutations of the response behave like a uniformly distributed vector on
the sphere, we generated a $100 \times 10,000$ data matrix $\bX$ with independent standard Gaussian
entries.  We then computed $\lambda_0(\bz_i)$ for each of 10,000 uniformly distributed vectors on $S^{n-1}$,
and compared the density of the resulting values to the values $\lambda_0(\by_{\pi_l})$ from 10,000 permutations
of a normalized response $\by$ in two settings: a sparse setting with $s = 10$ true variables and a signal to noise ratio 
(SNR, defined in (\ref{eq:SNR})) equal to 2; and
a less sparse setting with $s = 1000$ true variables and SNR equal to 2.   
The results, shown in Figure \ref{fig:unif},
demonstrate good agreement in both cases with the penalties derived from independent uniform responses.  Simulation results for data matrices $\bX$ with
different correlation structures (not shown) were similar.

\section{Simulations}
\label{sec:sims}

We performed a simulation study in which the proposed permutation selection procedure and competing methods were used to
select the penalty parameter of the LASSO.  Competing methods were BIC, 10-fold CV, and a procedure based on the 
covariance test of \citet{Lockhart2013}.
Simulated data sets differed in their dependence and level of sparsity (number of truly active predictors).
Both linear (Gaussian) and logistic (binomial) settings were considered.

\subsection{Predictor Matrix Generation}

We simulated predictor matrices with independent, multivariate Gaussian rows (samples).  In particular, the $n$ rows of $\bX$ 
were generated as independent samples from a $\mathcal{N}_p (\mathbf{0}, \boldsymbol{\Sigma})$ distribution.
Several structures for the covariance matrix $\boldsymbol{\Sigma}$ were considered.
\begin{enumerate}[(A)]
	\item Independent: $\boldsymbol{\Sigma} = \mathbf{I}_{p}$;
	\item Block correlation structure: $\sigma_{ij} = 0.5$ if $i \, \mathrm{mod} \, 10 = j \, \mathrm{mod} \, 10$ and $\sigma_{ij} = 0$ otherwise;
	\item Fast decaying AR(1) correlations: $\sigma_{ij}= 0.9^{|i-j|}$;
	\item Slowly decaying AR(1) correlations: $\sigma_{ij}= 0.99^{|i-j|}$.
\end{enumerate}

\subsection{Response Generation}
Given the predictor matrix $\bX$ and a simulated effects vector $\boldsymbol{\beta}$ (for more details, see below), 
we modeled the response using the generalized linear model (GLM) in equation \ref{eq:glm}. 
Two GLM settings were considered: Gaussian (standard linear model) and Bernoulli (logistic regression).

\subsubsection{Linear Regression Model}
Gaussian regression responses were simulated based on the standard linear regression version 
of equation (\ref{eq:glm}), in which the link function $g(\cdot)$ is the identity. 
Given a set of $s$ true variables selected uniformly at random, we generated the components of the 
associated effect vector $\boldsymbol{\beta}_s$ independently from a $\beta_i \sim U(0.25,1)$ distribution.
Letting $\mathbf{X}_s$ be the restriction of the predictor matrix to the columns of the $s$ selected variables,
the response vector $\by$ was generated as
\begin{equation}
\label{gausslm}
	\by \ = \  \mathbf{X}_s^\mathrm{T} \boldsymbol{\beta}_s + \boldsymbol{\epsilon}
\end{equation}
where $\boldsymbol{\epsilon} \sim \mathcal{N}(0, \sigma^2 \, \mathbf{I_n})$ is a vector of independent Gaussian errors.  
The variance $\sigma^2$ was selected to achieve a desired signal to noise ratio (SNR) given by
\begin{equation}
	\label{eq:SNR}
	\mathrm{SNR} \ = \  \sigma^{-2} \, \boldsymbol{\beta}_s^\mathrm{T} \, \text{Var}(\mathbf{X}_s) \, \boldsymbol{\beta}_s \, .
\end{equation}
One hundred simulations were performed for each choice of the following simulation parameters (64 in total):
\begin{itemize}

\item sample correlation structures (A)-(D);

\item SNR $= 0.5, 2$;

\item number of subjects $n = 200, 1000$; 

\item number of true variables $s = 1, 5, 10, 20$. 

\end{itemize}
In each case, the number of variables $p$ was equal to 500.

\subsubsection{Logistic Regression Model}

Logistic regression responses were simulated based on Bernoulli draws from the logistic version of the 
generalized linear model (\ref{eq:glm}) with link function
\begin{equation}
	\text{logit}(q) \ = \ \text{log}\left( \frac{q}{1-q} \right) \,,
\end{equation} 
where $q$ is the probability that a given subject is a case. 
In particular, for each sample $i$ we generated a value
\begin{equation}
	q_i \ = \ \mu + \bx_{i,s}^\mathrm{T} \, \boldsymbol{\beta}_s \,,
\end{equation}
where $\mu= \mathbf{1}^\mathrm{T} (\mathbf{X}_s^\mathrm{T} \, \boldsymbol{\beta}_s)$ is the intercept 
needed to obtain expected balance in the number of cases and controls, 
$\bx_{\cdot,s}$ are the $s$ predictors having true effects, and $\boldsymbol{\beta}_s$ are the effects,  
selected independently with $\exp\{\beta_{sj}\}\sim\mathrm{N}(\mu_\beta,0.02^2)$. 
The phenotype of sample $i$ was drawn from a $\mathrm{Bernoulli}(q_i)$ distribution. 
The parameter $\mu_\beta$ was used to control the signal level in each of the high and low dimensional data settings.
For the low dimensional setting, we selected $\mu_\beta$ as $\text{log}(1.15)$ and $\text{log}(1.35)$ for 
low and high signal levels, respectively.
In the high dimensional setting, we selected $\mu_\beta$ as $\text{log}(1.75)$ and $\text{log}(2.5)$ for low 
and high signal levels, respectively.
For each signal level, one hundred simulations were performed for each combination of correlation structure, 
sample size $n$, and number of true variables $s$ used in the Gaussian setting.

\subsection{Competing $\lambda$-selection methods}

\subsubsection{Bayesian Information Criterion (BIC)}

A standard version of BIC was implemented, selecting penalty parameter $\lambda_\text{BIC}$ via the relation
\begin{equation}
\lambda_\text{BIC} 
\ = \ 
\underset{\lambda}{\text{argmin}}  
\left\{ -2 \ell(\hat{\boldsymbol{\beta}}(\lambda: \by, \bX) : \by, \bX) \, + \, \mbox{df}(\lambda) \, \text{log}(n) \, \right\}, 
\end{equation}
where $\mbox{df}(\lambda) = | S_0(\lambda) |$ is the number of non-zero coefficients in the
coefficient vector $\hat{\boldsymbol{\beta}} (\lambda : \by, \bX)$.  The resulting variable set is 
$S_0(\lambda_\text{BIC})$

\subsubsection{Cross Validation (CV)}

Cross-validation was implemented using the cv.glmnet($\cdot$) function from the R-package r/glmnet, as 
described in \citep{Friedman2010}. Specifically, we selected the value of $\lambda$ minimizing the K-fold 
cross validation error 
\begin{equation}
\lambda_\text{CV} 
\ = \ 
\argmin_\lambda \left\{ 
\sum_{k=1}^{K} \sum_{i \in V_k} 
\left[y_i \, - \, \bx_i^T \hat{\boldsymbol{\beta}}^{(-k)}(\lambda : \by^{(-k)},  \bX^{(-k)} ) \right]^2 \right\} .
\end{equation}
Here $V_1,\ldots, V_K$ is a randomly chosen partition of $\{1,\ldots,n\}$ into groups of size 
$\lfloor n/K \rfloor$ or $\lceil n / K \rceil$, and 
the superscript $(-k)$ indicates that the subjects in $V_k$ have been excluded.
The resulting variable set is $S_0(\lambda_\text{CV})$.
We report results based on the value $K=10$.  The results for $K=3$ and $K=n$ 
(also referred to as the jackknife) were similar.

\subsubsection{Covariance Test}
The covariance test of \citet{Lockhart2013} provides a list of p-values corresponding to points 
in the LASSO path where a variable is added to the model. 
In principle, these p-values can be used to select a value of the penalty parameter;  
\citeauthor{Lockhart2013}\ left the 
specification of such a procedure for future work.  
Here we propose a simple procedure to select a penalty parameter $\lambda_\text{CT}$ using the p-values 
from the covariance test.  Let $p_{r}$ be the p-value produced by the covariance test for the $r$th change
in the LASSO path, and let
\begin{equation}
r^* \ = \ \max\{ r : p_{r} \leq \alpha \},
\end{equation}
where $\alpha$ is a fixed significance level, which we take to be $0.05$ in what follows.  
Define $\lambda_\text{CT}$ to be the value of $\lambda$ at which the $r^*$th change in
the LASSO path takes place;  the resulting variable set is then $S_0(\lambda_\text{CT})$.
In the case that no p-values are less then $\alpha$, we define $S_0(\lambda_\text{CT}) = \emptyset$.

Use of $\lambda_\text{CT}$ is limited to settings in which the covariance test can be applied.  At present the covariance
test cannot be applied in high dimensional ($p > n$) linear regression settings when the error variance is unknown.  
Computation of p-values for the covariance test requires knowledge of the LASSO path, in particular, values of 
$\lambda$ where model changes take place.  Use of the LARS algorithm for this purpose can be problematic: in several
of the simulated examples, the LARS algorithm encountered matrix singularities for the active set of variables, 
and we were unable to exactly identify $r^*$.  
In such cases, we identify $r^*$ based on the set p-values available from the portion of the 
LARS path that was successfully fit.
We note that current methods for LASSO fitting such as r/glmnet 
assess the LASSO path at a grid of $\lambda$ values 
(e.g., 100 values between $\lambda_0$ and $\epsilon\lambda_0$ for epsilon close to 0). 
This grid evaluation allows all other selection methods considered here to be fully evaluated; 
but as such grid based procedures are unable to locate the exact change points on the path, 
they can not be used for the covariance test.

\subsection{Simulation Results}

Across our simulations, the relationships between penalty selection methods were relatively consistent,
with overall performance depending primarily on the complexity of the chosen 
parameters (stronger dependence, lower SNR, fewer samples, and more active variables).
Below we describe the simulation results using the block correlation structure (covariance structure B), 
which are representative of our overall findings.
Figures showing simulation results under other settings can be found in Appendix \ref{sec:append}.

\subsubsection{Gaussian Simulation Results}

For the classical ($n > p$) setting we were able to apply all four selection methods.
Application of the covariance test to high dimensional ($p > n$) problems when the 
error variance is unknown is still under investigation.
Rather than exclude $\lambda_\text{CT}$ from the high dimensional simulations, 
we evaluated $\lambda_\text{CT}$ using the simulated error variances.
Thus the covariance test is given a substantial advantage in the high dimensional settings, and
the results from these settings should be interpreted accordingly.

\begin{figure}[ht!]
  	\centering
		\includegraphics{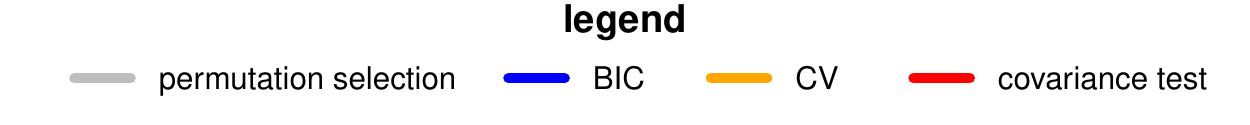}\\
		\includegraphics[width=.45\linewidth]{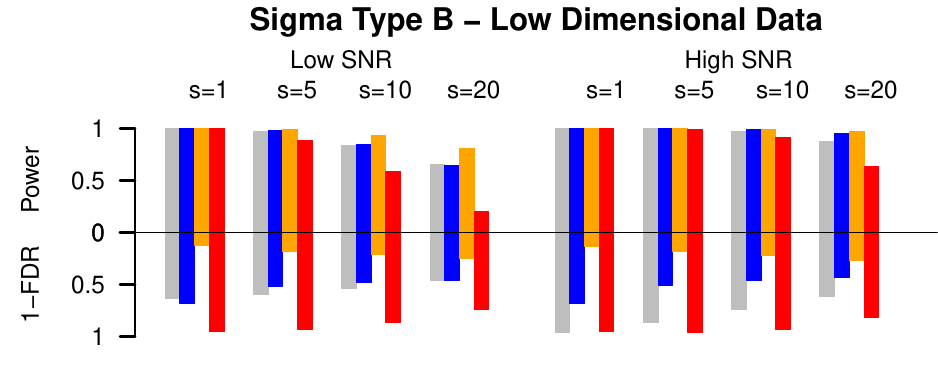}	\includegraphics[width=.45\linewidth]{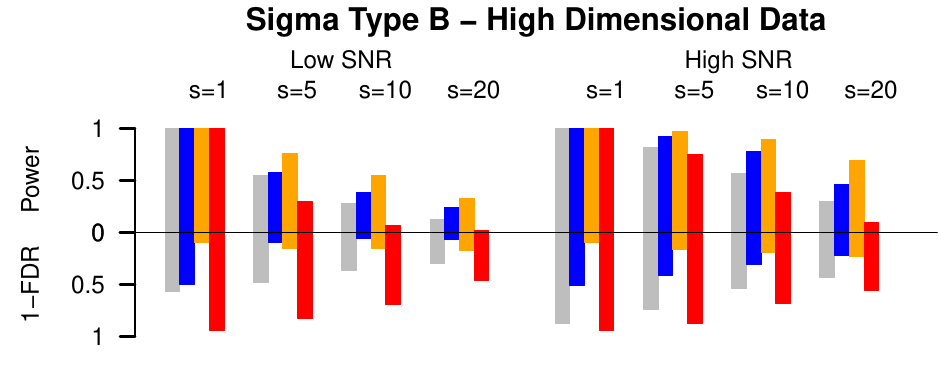}\\
	\caption{Bar plots of power and false discovery rate (FDR) for selected gaussian regression simulation settings. s indicates the number of true signals. }
  	\label{fig:bar_g}
\end{figure}

Figure \ref{fig:bar_g} shows the average power (number of true discoveries divided by total
number of active variables) and the average false discovery rate (number of false
discoveries divided by total number of discoveries) for high and low SNR regimes,
and for different numbers of true variables $s$.
Examining the results from the low dimensional data setting (left), we observe a consistent 
relationship between power and FDR.  
Specifically, we find that CV has the highest power along with the largest FDR.  
Permutation selection and BIC perform similarly, with BIC having slightly increased power and FDR.  
The covariance test is the most conservative of those considered, having both lower power and lower FDR 
than the other methods.  
In the high dimensional setting (right), we observe a similar relationship between power and FDR.  
The principal difference is that BIC and permutation selection have greater differences in 
power and FDR in the high dimensional setting.  Specifically, while BIC's advantage in power over 
permutation selection is slightly larger in the high dimensional setting, it comes with a greater 
increase in FDR than that observed in the low dimensional setting.

\subsubsection{Logistic Simulation Results}

\begin{figure}[ht!]
  	\centering
	\includegraphics{legend.pdf}\\
		\includegraphics[width=.45\linewidth]{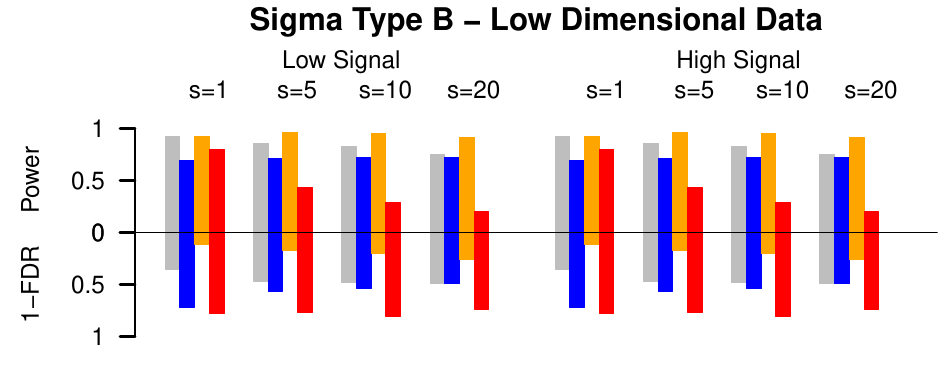}	\includegraphics[width=.45\linewidth]{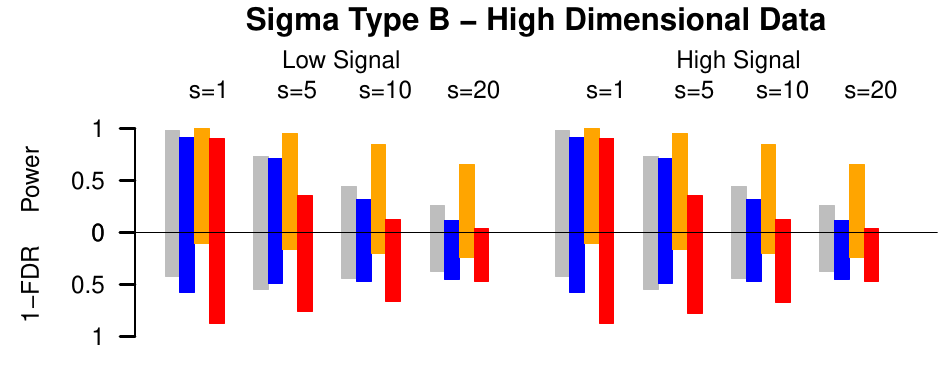}\\
	\caption{Bar plots of power and false positive rate (FPR) for selected logistic regression simulation setting. s indicates the number of true signals.}
  	\label{fig:bar_l}
\end{figure}

Figure \ref{fig:bar_l} summarizes the results of the logistic simulations.
(The covariance test is applicable to logistic regression in both low and high dimensional settings.)
Examining the low dimensional data setting (left), we observe that CV has the highest power, but also the highest FDR.  
Permutation selection is generally comparable, or superior, to BIC.  The covariance test performs best for $s=1$; in the low
SNR regime its performance degrades as $s$ increases.  Overall, the covariance test is more conservative, with lower
power and lower FDR than the other methods.  Similar trends are seen with the high dimensional data.

\section{Real Data Analysis}
\label{sec:realdata}

Here we apply all four considered penalty parameter selection methods on several real data sets. The data sets examined are:

\begin{itemize}
	\item Internet-Ad data \citep{Kushmerick_1999}: A document classification problem consisting of mostly binary features. The response is binary, and indicates whether the document is an advertisement. Only 1.2\% of the values in the predictor matrix are nonzero. The data consist of 2358 observations of 4290 variables.  (This data is publicly available at http://www.stanford.edu/hastie/glmnet/)
	\item The Cancer Genome Atlas Network (TCGA) breast cancer data \citep{TCGAbreast}: Tumor and germline DNA samples were obtained from 825 patients. Rsem gene expression values were normalized to the upper quartile of the total counts.  The data was then log2 transformed and genes were median centered.  Genes were filtered such that at least 70\% of samples had a value, and then imputation was performed. (This data is available from the TCGA data portal at  https://tcga-data.nci.nih.gov/tcga/) We examine the problem of distinguishing 371 Luminal A cancer samples from 170 Luminal B cancer samples based on the expression values of the 17,007 genes that remained after quality control. 
\end{itemize}

We consider the computational time of each fitting procedure and,
although our primary focus is variable selection, we also examine the predictive ability of the the fitted 
models. 
For each analysis, we randomly split the data set into a training and a test set. 
The training set consists of two thirds of the data, and the remaining third is used for testing.   
The LASSO was applied to the training data with the penalty parameter selected by each method,
and predictive performance was evaluated on the test set. 
Random splitting of each data set was repeated 10 times.

\subsection{Internet Ad Data Analysis}
Table \ref{tab:realdata_ad} summarizes the sizes of the selected models, 
the percent of the test set misclassified, 
and computational time for each penalty selection method.  
When applying the covariance test procedure, we found that collinearity in the data created problems, and  
we only had access to the first 15 steps in the LASSO path (with some variability for different splits).
This has two consequences: (i) the model resulting from the covariance test may be larger than 
if we had access to the full path; and (ii) the reported computation time for the covariance test is 
significantly reduced.  
Examining how often the different procedures misclassified the test set, we found that 
the prediction based CV method had the best performance, followed by BIC and permutation
selection.  The more conservative covariance test had the highest misclassification rate.
As expected, the simple permutation selection procedure was substantially faster than all 
other methods.

\begin{table}[ht!]
	
	\centering
	\begin{tabular} {l | c c c}
		\hline
		Method	&	Model Size	&	Percent Misclassified & CPU seconds 	\\
		\hline
		BIC		&	35.6 (2.01)			&	5.07 (0.36) 	& 12.08 (0.32)		\\
		covTest$^\star$	&	6 (0.88)	&	9.89 (0.84) 	& 18.57 (1.91)	\\
		CV		&	123.7 (6.67)		&	3.46 (0.23) 	& 34.35 (0.95)	\\
		Permutation Selection	&	26.5 (2.13)			&	6.45 (0.28)		& 3.69 (0.04)	\\
		\hline		
	\end{tabular}
	\caption{Means and standard errors of the model sizes, percent of test set misclassified, and computation times from 10 random splits of the Internet Ad data. $^\star$ indicates that the exact model may be larger, but LARS was unable to fit the entire path, which also reduced computation time. }
	\label{tab:realdata_ad}
\end{table}

\subsection{TCGA Data Analysis}
Table \ref{tab:realdata_lum} summarizes the sizes of the selected models, the percent of the test set misclassified, and computational times for each penalty selection method.  
Collinearity issues arose again in our application of the covariance test, and we therefore 
limited our analysis to the first 100 steps in the LARS path; 
as with the Internet Ad data, this significantly lowered the computational time for the covariance test.
Examining the percent of the test set misclassified by the models, we find that CV has the best 
performance, followed by permutation selection and BIC.
The covariance test has a significantly higher misclassification rate than the other methods. 
BIC and permutation selection have similar computational times, closely followed by CV.  In this
analysis the covariance test was significantly (more than one order of magnitude) slower than the other methods.

\begin{table}[ht!]
	
	\centering
	\begin{tabular} {l | c c c}
		\hline
		Method			&	Model Size			&	Percent Misclassified 	&	CPU seconds \\
		\hline
		BIC				&	18.5 (2.62)			&	14.61 (1.28)			&	9.43 (0.12)		\\
		covTest$^\star$	&	8.4 (5.15)			&	26.33 (2.47) 			&	601.05 (6.72)	\\
		CV				&	85.8 (4.12)			&	8.05 (0.68) 			&	14.15 (0.19)	\\
		Permutation Selection			&	22.9 (0.81)			&	13.00 (0.51) 			&	10.75 (0.26)	\\
		\hline		
	\end{tabular}
	\caption{  Means and standard errors of the model sizes, percent of test set misclassified, and computation times from 10 random splits of the TCGA luminal subsets data. $^\star$ indicates that the exact model may be larger, but LARS was unable to fit the entire path, which also reduces computation time. }
	\label{tab:realdata_lum}
\end{table}

\section{Summary}
\label{sec:sum}
From the simulated and real data analyses, several trends are apparent.  
As expected, the prediction based cross validation method tends to select larger models from the LASSO sample path.  These models have high power, but relatively large false discovery rate; other penalty selection methods are preferred when variable selection is the primary goal of the analysis.
In contrast with cross validation, the covariance test tends to select smaller models, with lower power and lower false discover rate.  Permutation selection and BIC lie somewhere between cross validation and the covariance test, favoring moderate sized models and a more balanced tradeoff between power and false discovery rate.  

The covariance test method performs best, often beating other methods, when the number $s$ of true variables is small, but its performance drops off as the number of true variables increases.  This may reflect the difficulty of testing changes
in the LASSO path when conditioning on more complex models containing a mix of true and spurious variables.
At present, application of the covariance test to high 
dimensional ($p > n$) linear models is limited 
to settings in which the error variance is known or can be accurately estimated.   
In our high dimensional linear model simulations, we provided the true error variance to the 
covariance test procedure.
We note that in cases where the number of true variables is known to be very small, simple predictor-by-predictor selection methods may outperform the LASSO, regardless of how the penalty parameter is selected. 

The permutation selection method is straightforward in its implementation and interpretation.  
In our simulations, permutation selection is comparable or superior to BIC, and is generally competitive with the covariance test, the former having the advantage for large values of $s$, the latter for small values of $s$. 
In the real data analyses, permutation selection and BIC were roughly comparable in terms of model size and predictive performance.  
The variables selected by the covariance test were also selected by BIC and permutation selection. 
As it tends to produce smaller models, the covariance test procedure had relatively poor 
predictive performance.  

Cross validation, BIC, and the covariance test require computation of the full LASSO path on the given data.  By contrast, permutation selection requires only the initial point of the LASSO sample path, but under multiple permutations of the response.  On balance,
the efficiency with which the initial point of the LASSO path can be identified makes permutation selection faster, sometimes by an
order of magnitude, than the other methods considered.

\section{Discussion}
\label{sec:disc}
When selecting models using penalized regression methods such as the LASSO, selection of the penalty parameter 
is an important part of the analysis process.
Selection is often done with the goal of prediction or variable selection.
The most common methods for selecting the penalty parameter are cross validation (CV) and 
the Bayesian information criterion (BIC).
In this paper we introduced a simple permutation based selection procedure that is intended for situations where
variable selection is a primary goal of model fitting.
Permutation selection chooses a model whose variables have a joint relationship with the response 
that is stronger than joint relationships occurring at random, for permuted versions of the response. 

Permutation selection is fast and yields interpretable models.
In our simulations and real data analyses, permutation selection was comparable to BIC and competitive
with the covariance test procedure.  At present, application of the covariance test to high 
dimensional ($p > n$) linear models is limited 
to settings in which the error variance is known or can be accurately estimated.   
Although we have focused here on the LASSO, permutation selection
is applicable in principle to other penalized regression methods such as SCAD.

We have considered settings in which all variables are subject to penalization.
In some cases it is natural to consider models in which select variables (for example, predictors
known to affect the response) are unpenalized.  
Although CV and BIC can be extended to this situation,
permutation selection does not immediately apply, as permutation of 
the response will nullify its relationship between penalized and unpenalized variables alike.    
We leave to future work a detailed investigation of how permutation selection can be applied 
to applications with penalized and unpenalized variables.

Both permutation selection and cross validation make use of randomization, which leads to variability in the 
selection of the penalty parameter.  For permutation selection this variability is controlled by considering the
median of the null penalties across a moderate number (100) of permutations.  
In most applications with moderate
or small sample sizes, the randomness of the sample itself is a more significant 
source of variability, one that is often not accounted for in applications of the LASSO, regardless of how the
penalty parameter is selected.
Recently, a number of resampling based methods have been proposed to address the stability 
of models with respect to variability of the sample.  We refer the interested reader to 
\citet{Valdar2009}, \citet{Meinshausen10}, and \citet{ValdarSabourin12} for more details.

\newpage

\appendix

\section{Full Results Plots}
\label{sec:append}

\subsection{Gaussian Model Results}

\begin{figure}[ht!]
  \centering
  	\includegraphics{legend.pdf}\\
	\includegraphics[width=.45\linewidth]{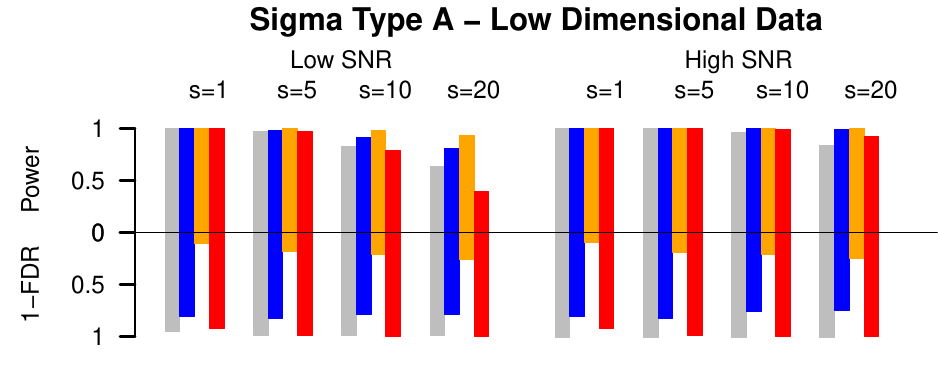}	\includegraphics[width=.45\linewidth]{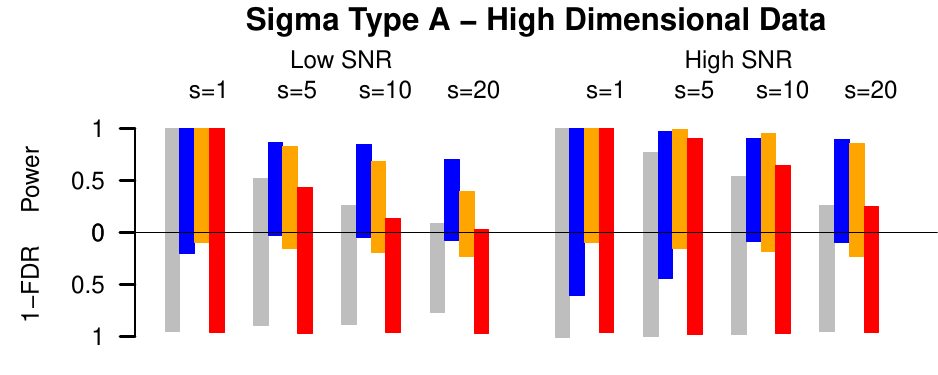}\\
	\includegraphics[width=.45\linewidth]{new2_1-FDR_resultsBar_B_LowdimFDR.pdf}	\includegraphics[width=.45\linewidth]{new2_1-FDR_resultsBar_B_HighdimFDR.pdf}\\
	\includegraphics[width=.45\linewidth]{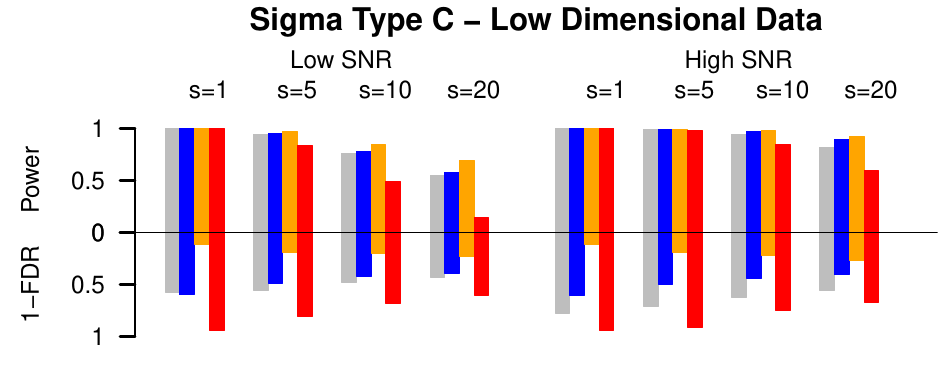}	\includegraphics[width=.45\linewidth]{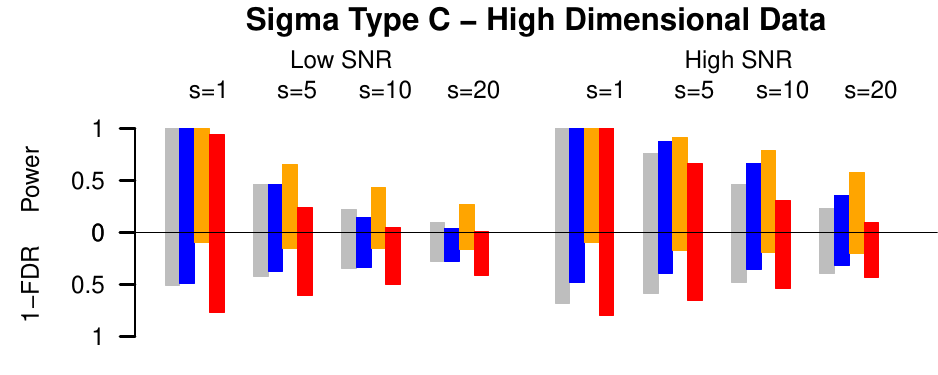}\\	
	\includegraphics[width=.45\linewidth]{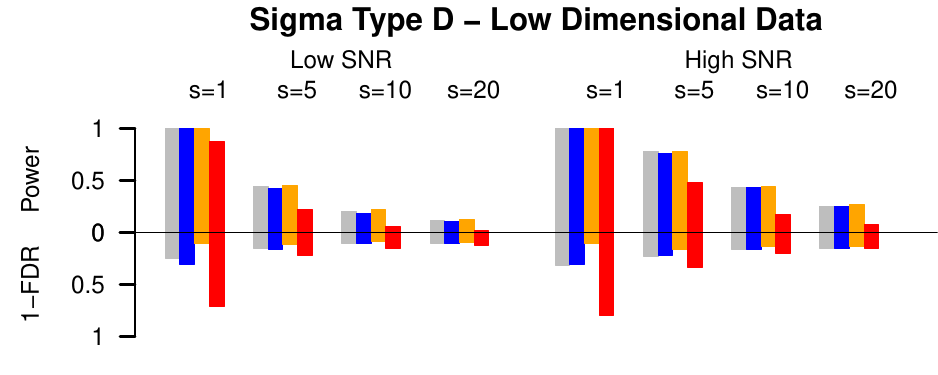}	\includegraphics[width=.45\linewidth]{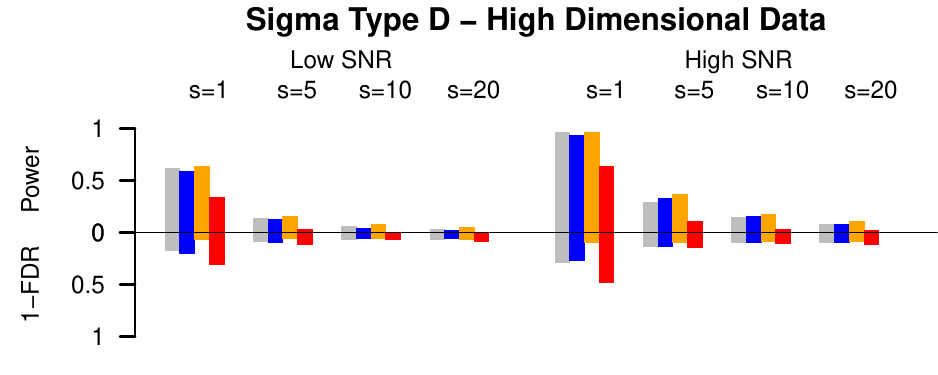}\\	
  \caption{Bar plots of power and $1 -$ false discovery rate (FDR) for each gaussian regression simulation setting. }
  \label{fig:allbar_g.fdr}
\end{figure}

\newpage

\subsection{Logistic Model Results}
\begin{figure}[ht!]
  \centering
  	\includegraphics{legend.pdf}\\
	\includegraphics[width=.45\linewidth]{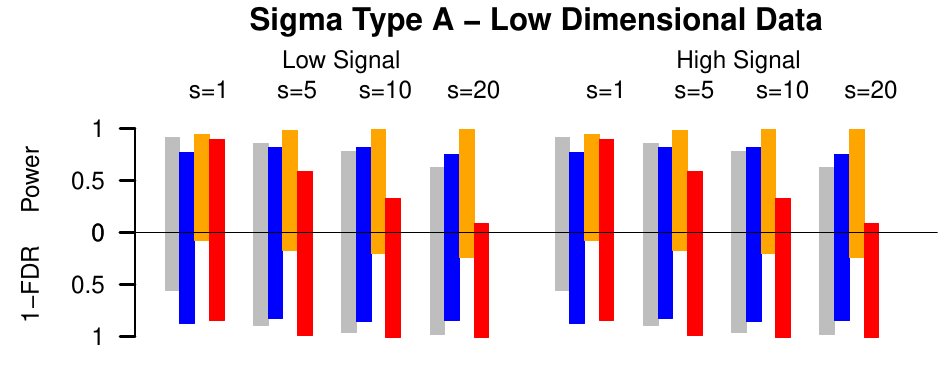}	\includegraphics[width=.45\linewidth]{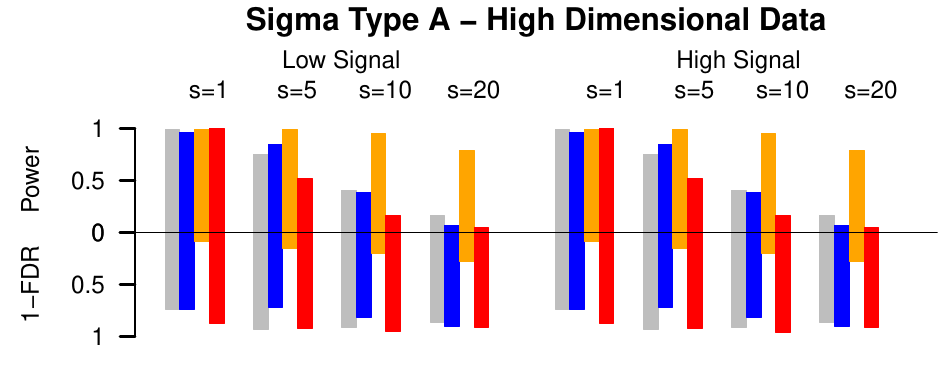}\\
	\includegraphics[width=.45\linewidth]{resultsBar_1-FDR_B_LowdimFDR_logistic.pdf}	\includegraphics[width=.45\linewidth]{resultsBar_1-FDR_B_HighdimFDR_logistic.pdf}\\
	\includegraphics[width=.45\linewidth]{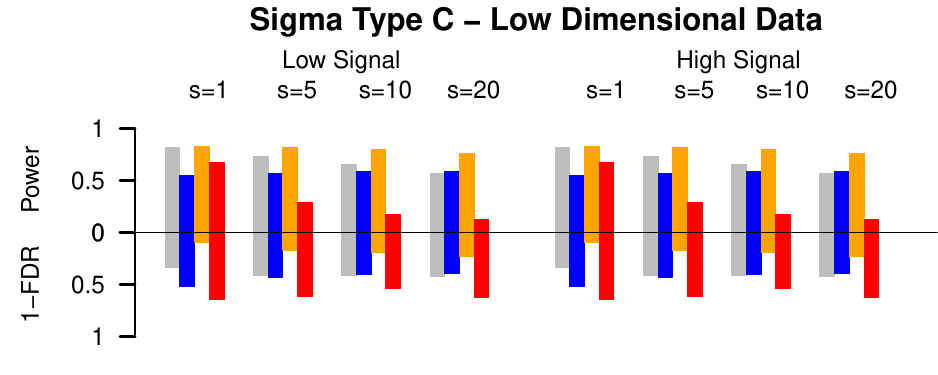}	\includegraphics[width=.45\linewidth]{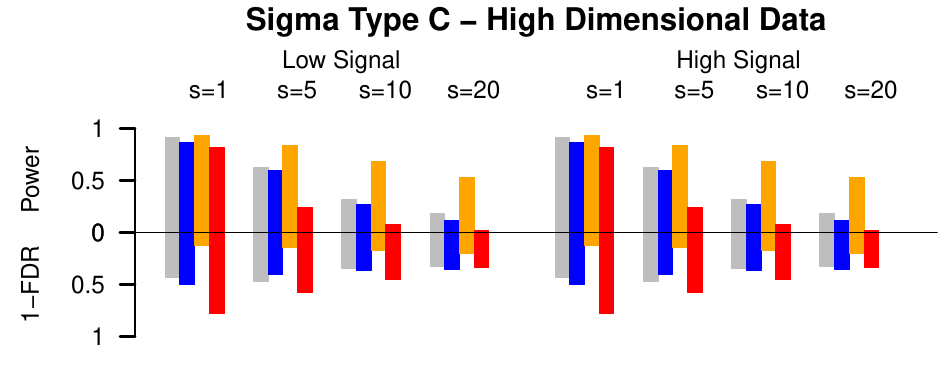}\\	
	\includegraphics[width=.45\linewidth]{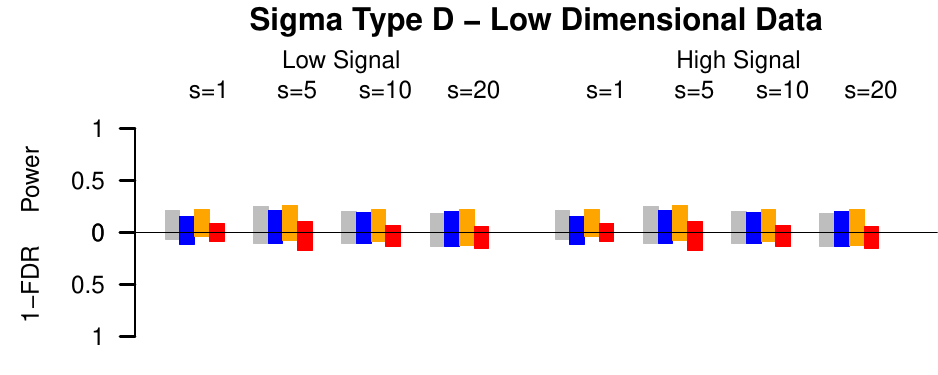}	\includegraphics[width=.45\linewidth]{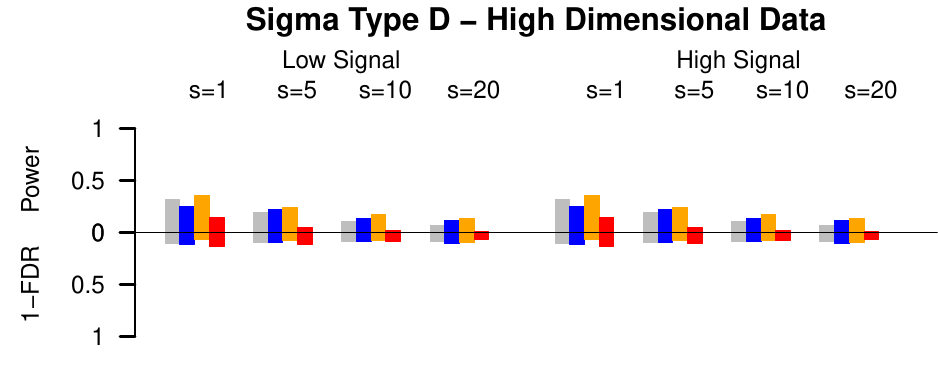}\\	
  \caption{Bar plots of power and $1 -$ false discovery rate (FDR) for each logistic regression simulation setting. }
  \label{fig:allbar_l}
\end{figure}

\bibliography{LambdaPerm}
  \bibliographystyle{genepi5auth}
  
\end{document}